\documentclass[10pt,twocolumn,letterpaper]{article}

\usepackage[pagenumbers]{cvpr} 

\usepackage[accsupp]{axessibility}  

\usepackage{makecell}
\usepackage{xcolor}
\usepackage{sectionbox}
\usepackage{tcolorbox}

\definecolor{cvprblue}{rgb}{0.21,0.49,0.74}
\usepackage[pagebackref,breaklinks,colorlinks,allcolors=cvprblue]{hyperref}

\def\paperID{****} 
\def\confName{CVPR}
\def\confYear{2026}

\title{FLARE: A Failure-Aware Framework for Autonomous Correction and Recovery in Visual-Language Robotic Manipulation}

\author{
Ganlong Zhao\textsuperscript{1,2}
\
Zijia Tang\textsuperscript{3}
\
Xingping Chen\textsuperscript{4}
\
Zhanghui Kuang\textsuperscript{5}
\
Ye Tian\textsuperscript{6}
\
Guanbin Li\textsuperscript{4,7,8$^*$}
\\
\textsuperscript{1}The Chinese University of Hong Kong 
\quad \textsuperscript{2}Centre for Perceptual and Interactive Intelligence
\\
\quad \textsuperscript{3}Duke University
\quad \textsuperscript{4}Sun Yat-sen University
\quad \textsuperscript{5}TengenX
\quad \textsuperscript{6}Tencent Robotics X\\
\quad \textsuperscript{7}Shenzhen Loop Area Institute 
\quad \textsuperscript{8}Guangdong Key Laboratory of Big Data Analysis and Processing
\\
{\tt\small glzhao@cpii.hk, liguanbin@mail.sysu.edu.cn 
}
}

\begin{document}
\maketitle

\def\thefootnote{*}\footnotetext{Corresponding author is Guanbin Li.}

\begin{abstract}

Vision-Language-Action Models~(VLAs) have demonstrated significant promise in generalizing to complex, long-horizon robotic manipulation tasks. However, their performance remains brittle, as they are typically trained on trajectory-monotonic, failure-free demonstrations. This reliance on ``perfect" data leaves them unable to recover from common execution errors, such as a missed grasp, a dropped object, or an unexpected collision. 
In this paper, we propose FLARE, a novel framework that endows VLAs with robust error recovery capabilities through a ``Retry" and ``Reset" paradigm. 
First, we introduce a ``Retry" mechanism by injecting perturbation and bridging segments that decouple robot pose from environment state into demonstrations, enabling the policy to autonomously handle execution deviations.
Second, to address critical, state-breaking (OOD) failures, we introduce a ``Reset" pipeline. We leverage an MLLM for offline failure analysis to automatically identify OOD states from execution videos. This analysis enables the efficient, targeted collection of a small library of object-centric ``Reset" skills, which are trained to restore the environment to a task-valid state.
Our full framework integrates these learned policies. At inference, an online MLLM monitor arbitrates between task execution and ``Reset" skills. Experiments on challenging, contact-rich manipulation tasks show our approach significantly improves task success and robustness.

\end{abstract}    
\section{Introduction}
Vision-Language-Action (VLA) models have recently emerged as a promising paradigm for Embodied Artificial Intelligence, integrating perception, language, and control to enable robots to execute complex, long-horizon tasks in open-world environments~\cite{black2024pi_0,zitkovich2023rt,kim2025openvla,intelligence2025pi_,ma2024survey,zhao2024over,zhao2025aerial}. Despite impressive advances—such as $\pi_0$~\cite{black2024pi_0} and OpenVLA~\cite{kim2025openvla}—current systems remain notably brittle: small perturbations, unexpected object contacts, or slight execution deviations can cause irreversible task failures. Unlike humans, VLAs lack an intrinsic ability for continuous self-correction. Achieving true robotic autonomy therefore requires not only skillful task execution but also a robust and generalizable mechanism for recovery after failure.

\begin{figure}
    \centering
    \includegraphics[width=0.9\linewidth]{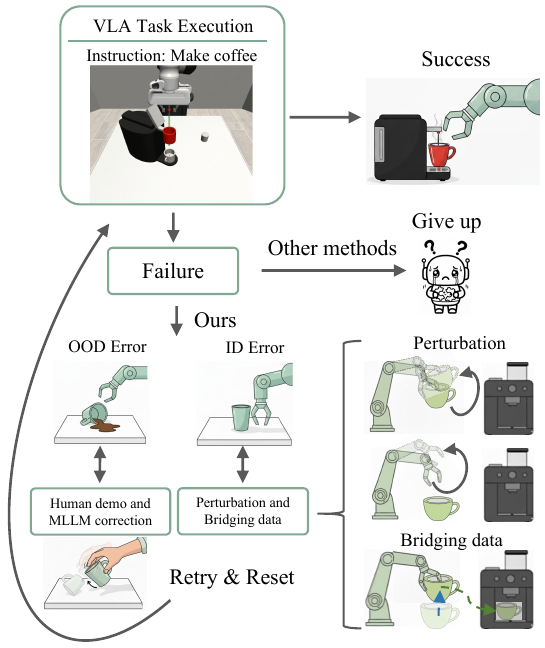}
    \caption{FLARE: Failure-Aware Resilience in VLA. Previous methods are brittle, failing from minor perturbations (ID errors) or catastrophic states (OOD errors), and do not leverage the VLA's inherent generative capability for continuous self-correction. Our FLARE framework introduces an ID/OOD error taxonomy to guide a dual Retry/Reset paradigm. We tackle ID errors via built-in Retry robustness and resolve OOD errors with MLLM-bootstrapped Reset skills, achieving robust, resilient autonomy.}
    \label{fig:teaser}
\end{figure}

While previous research attributes brittleness to model architecture or control policy limitations~\cite{guo2025robustness,fei2025libero,zhang2025robustvla}, we identify a deeper origin in the data regime itself. Human demonstrations are costly to collect, resulting in datasets that are sparse, success-biased, and lacking in trajectory diversity. Even data recomposition methods such as MimicGen~\cite{mandlekarmimicgen} expand dataset volume but fail to address two fundamental issues.
First, both original and recombined demonstrations are trajectory-monotonic: the robot consistently follows a narrow manifold of poses correlated with task progress. Consequently, the model learns spurious associations between its own configuration and task success. When encountering a valid yet unseen configuration—an in-distribution (ID) error—the agent cannot resume appropriately.
Second, these datasets are success-centric and contain few, if any, examples of recovery. Catastrophic failures (e.g., a toppled cup) drive the environment into an out-of-distribution (OOD) error state, for which the agent has never observed corrective behavior. Since standard augmentations recombine only successful data, they cannot synthesize recovery trajectories, leaving the system unable to restore a valid task state.

Recent studies have explored several approaches for robotic self-correction. One direction leverages Multimodal Large Language Models~(MLLMs) to provide semantic feedback~\cite{liu2025coherent,wang2023describe,xia2024kinematic,xiong2024autonomous}; these systems can recognize high-level failures but typically depend on a fixed skill library, limiting low-level adaptability. Another line of research employs reinforcement~\cite{lu2024koi,schulman2017proximal,yokoyama2023asc,horgan2018distributed} or instruction-based learning~\cite{xia2025phoenix,yang2025fpc}, which struggles with sample efficiency and generalization in long-horizon settings. A unified, data-centric mechanism capable of enabling self-recovery across both ID and OOD error regimes remains absent.

To this end, we propose FLARE, a Failure-Aware Retry/Reset framework designed to transform brittle VLAs into resilient embodied agents~(Fig.~\ref{fig:teaser}). At its foundation lies an ID/OOD error taxonomy and two complementary recovery mechanisms.
Retry addresses ID errors by systematically decoupling robot pose from environment state. We introduce a perturbation-bridging augmentation strategy that injects random pose perturbations between task segments, followed by a bridging segments that reconnects them. This process teaches the VLA to execute the correct next step from diverse configurations, providing built-in retry robustness.
Reset tackles OOD errors through efficient skill bootstrapping. An offline MLLM (e.g., Gemini~\cite{team2023gemini}) serves as a failure analyst, parsing failure videos to determine which object requires resetting. This analysis bootstraps the collection of a compact set of object-centric demonstrations (e.g., ``un-topple the cup''), which are further enhanced via the same perturbation-bridging augmentation. These reset behaviors restore the environment to a stable configuration, allowing task resumption.

All data—original, retry-augmented, and reset-augmented—are integrated to train a unified VLA system. This system equips the agent with a comprehensive repertoire of skills, where task execution and environment recovery are seamlessly invoked through distinct language prompts.
During deployment, an online MLLM monitor observes task execution and dynamically arbitrates between retry and reset modes. For ID errors, the VLA autonomously recovers through its learned robustness; for OOD errors, the monitor switches the prompt to the corresponding reset skill. Once the environment is restored, the system seamlessly resumes the main task. This retry–reset loop operationalizes robust autonomy, enabling resilience to both pose-level and environment-level disturbances.

In summary, our contributions are threefold:
(1) A unified Retry/Reset paradigm grounded in an ID/OOD error taxonomy that reframes robotic autonomy as resilience rather than perfection.
(2) A perturbation-bridging augmentation strategy that decouples robot pose from environment state, equipping VLAs with built-in retry robustness.
(3) An MLLM-driven dual-loop system in which the large language model serves both as an offline failure analyst for reset skill acquisition and as an online monitor for closed-loop recovery.

Overall, FLARE advances embodied intelligence from imitation-driven execution toward resilience-driven autonomy, enabling agents not only to act but to recover.
\section{Related Works}

\subsection{Robotic Self-correction Systems}
Research on robotic self-correction generally falls into three categories. First, MLLM-based approaches treat large multimodal models as high-level failure analysts~\cite{gupta2025perception,liu2025coherent,wang2023describe,xia2024kinematic,xiong2024autonomous,li2024self,lin2025failsafe,chen2024automating,xiong2024aic}. Systems such as REFLECT~\cite{liu2023reflect} can diagnose failures and replan, but their reliance on a fixed skill library restricts low-level adaptability. Second, reinforcement learning methods~\cite{lu2024koi,schulman2017proximal,yokoyama2023asc,horgan2018distributed,vats2024recoverychaining} aim to explicitly learn recovery behaviors; although frameworks like SeRO~\cite{kim2023sero} handle OOD states, they often face prohibitive sample-efficiency demands in real-world settings. Third, data-driven paradigms~\cite{xia2025phoenix,yang2025fpc} enhance robustness via large-scale augmentation. Phoenix~\cite{xia2025phoenix} connects semantic reflection to low-level correction through coarse motion instructions refined by human feedback, but its template-based prediction–correction pipeline limits generalization across tasks and environments.

\subsection{Vision-Language-Action Models} 
Recent advances in embodied AI have been driven by Vision-Language-Action (VLA) models~\cite{kawaharazuka2025vision,zhang2025pure,zhao2023learning}, which unify perception, language, and control within a single architecture. This paradigm was initiated by Google’s Robotic Transformer series: RT-1~\cite{brohan2023rt} scaled transformer-based policies to real-world control, and RT-2~\cite{zitkovich2023rt} further co-fine-tuned a large VLM on web and robotic data to enable semantic grounding.
Building on this foundation, OpenVLA~\cite{kim2025openvla} offers an open-source pipeline that fine-tunes modern VLMs~\cite{lillava,beyer2024paligemma,touvron2023llama,bai2023qwen,bai2025qwen2,driess2023palm} to output tokenized robot actions, exhibiting strong zero-shot generalization. Likewise, $\pi_0$~\cite{black2024pi_0} and $\pi_0$-FAST~\cite{pertsch2025fast} adopt a VLA Flow Model~\cite{lipman2022flow} built on pre-trained VLMs~\cite{beyer2024paligemma}, achieving competitive performance on long-horizon, multi-step manipulation tasks. Additional works explore more generalizable manipulation pipelines~\cite{zhao2023learning,chi2025diffusion,liang2023code,ghosh2024octo,o2024open}.
Despite their strong generalization within the distribution of training tasks, these models remain vulnerable to small perturbations, unexpected contacts, and minor execution drift—limitations noted in recent benchmarks~\cite{fei2025libero} and a key barrier to reliable open-world autonomy.

\section{Methodology}
\label{sec:method}

\begin{figure*}[t]
    \centering
    \includegraphics[width=\linewidth]{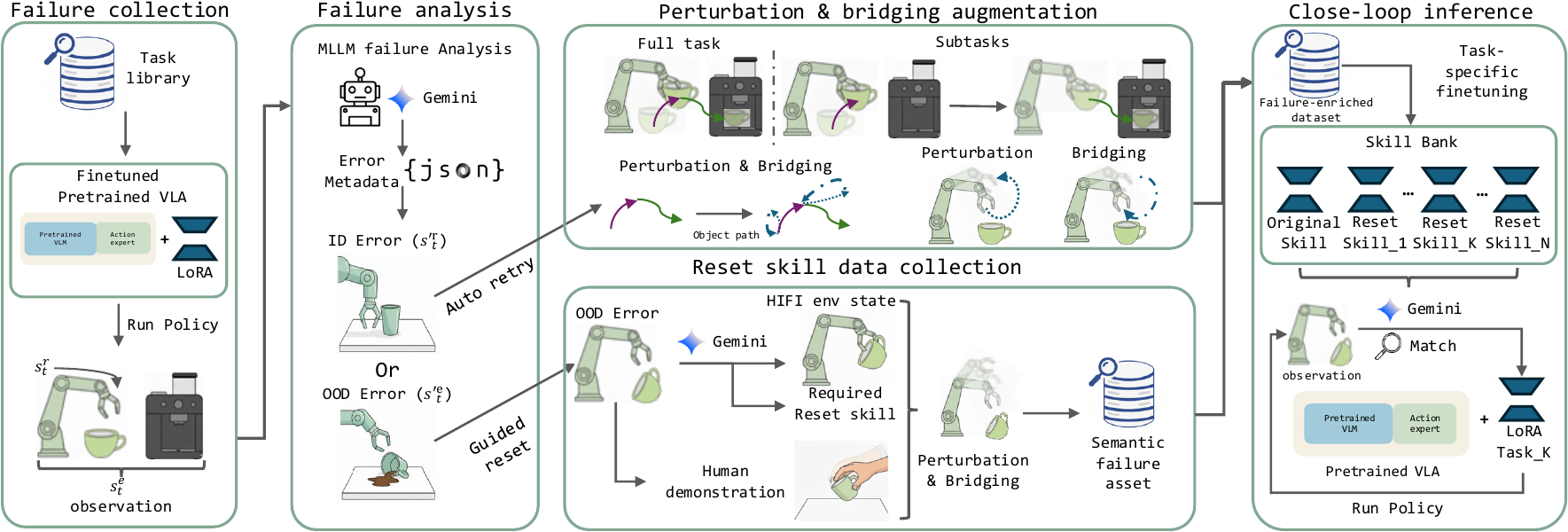}
    \caption{The overall framework of our method. We first collect the failure data with the VLA model trained with regular demonstrations. Then we perform the failure analysis with MLLM and formulate the failures with ID/OOD taxonomy. For ID errors, we perform the perturbation \& bridging augmentation to enhance the robustness and ``retry'' ability of VLA models. For OOD errors, we collect the object-centric reset skill data. All these data are utilized to obtained a skill bank for further close-loop inference.}
    \label{fig:framework}
\end{figure*}

Our goal is to endow VLAs with a robust, generalizable error recovery capability. We introduce the Retry/Reset framework, a unified approach built upon a taxonomy of failures as either In-Distribution (ID) or Out-of-Distribution (OOD) errors. ID errors represent states where the environment is still on a valid task trajectory, but the robot's pose is novel; these only require the robot to ``retry" the correct action from its new pose. OOD errors represent catastrophic failures (e.g., a toppled object) where the environment state is unrecoverable by the task policy, requiring a dedicated ``reset" skill. Our method provides a distinct solution for each case, training a unified VLA system to handle both~(Fig.~\ref{fig:framework}).

\subsection{Problem Formulation} \label{sec:formulation}

We model the VLA agent as a policy $\pi_{\theta}$. Following modern VLA architectures~\cite{kim2025openvla,black2024pi_0,intelligence2025pi_}, the policy is Markovian—lacking history—and predicts an action chunk $\mathbf{a}_t$ based on the current visual observation $o_t \in \mathcal{O}$ and language instruction $\mathcal{I}$. This policy, which outputs a distribution over action sequences $\mathbf{a}_t \in \mathcal{A}^K$ (where $K$ is the chunk length and $\mathcal{A}$ is the action space), is written as: $\mathbf{a}_t \sim \pi_{\theta}(\cdot | o_t, \mathcal{I})$. This policy is typically trained via imitation learning on a dataset of $N$ successful human demonstrations, $\mathcal{D}_{\text{demo}} = \{\tau_i\}_{i=1}^N$.

A core challenge arises from the nature of the observation $o_t$ and the demonstration data $\mathcal{D}_{\text{demo}}$. The true state of the world at time $t$, $s_t$, can be decomposed into two components: the environment state $s_t^e$ (e.g., the poses and states of all objects) and the robot state $s_t^r$ (e.g., the end-effector pose and gripper status). The visual observation $o_t \in \mathcal{O}$ is a rendering of the complete state $s_t$.

The ``trajectory-monotonic" nature of the demonstration dataset $\mathcal{D}_{\text{demo}}$ creates a strong, spurious correlation between $s_t^e$ and $s_t^r$. This is because human demonstrations are not exhaustive; they reflect implicit biases and habits. For example, a demonstrator might always grasp a cup by its handle or consistently approach from a similar angle, even if many other approaches are valid. This results in a highly-biased, low-variance conditional distribution $P(s_t^r | s_t^e, \mathcal{D}_{\text{demo}})$.

This correlation is a root cause of VLA brittleness. Trained on this biased data, $\pi_{\theta}$ overfits and fails to learn a disentangled representation. Instead of correctly inferring task progress primarily from the environment state $s_t^e$ (e.g., ``the cup is now grasped"), the policy incorrectly learns to associate task progress with the specific joint configuration $(s_t^e, s_t^r)$ seen in demonstrations. This leads to a critical failure: when a minor perturbation creates a state with a valid $s_t^e$ but a novel $s_t^r$, the policy incorrectly interprets this valid state as out-of-distribution and fails. For example, the robot's pose $s_t^r$ might be at a ``placing" position, causing the policy to infer the task is at the ``placing" stage, even if the environment state $s_t^e$ shows the object was never successfully grasped. This explains the paradoxical lack of ``retry" capability in existing VLAs.

We formalize this challenge by introducing a taxonomy of failure states. Let $\mathcal{S}_{\text{task}}^e$ be the set of all valid environment states along any successful task execution path.

Definition 1: In-Distribution~(ID) Error. An ID Error is a state $s_t = (s_t^e, s_t^r)$ where the environment state is valid ($s_t^e \in \mathcal{S}_{\text{task}}^e$), but the robot state $s_t^r$ is from a low-probability region of the demonstrated conditional distribution $P(s_t^r | s_t^e, \mathcal{D}_{\text{demo}})$. The task policy $\pi_{\theta}$ fails because it has overfitted to the $s_t^e$-$s_t^r$ correlation, even though the task is still recoverable from $s_t^e$. Recovery requires a ``Retry" capability.

Definition 2: Out-of-Distribution~(OOD) Error. An OOD Error is a state $s_t = (s_t^e, s_t^r)$ where the environment state itself is non-recoverable by the standard task policy ($s_t^e \notin \mathcal{S}_{\text{task}}^e$). For example, a toppled cup creates a state $s_t^e$ that no action from $\pi_{\theta}(\cdot | \mathcal{I}_{\text{task}})$ can salvage. Recovery requires a distinct ``Reset" skill.

Our goal is to learn a robust, unified policy system that can handle both error types. This system must:
\begin{enumerate}
\item Possess an innate ``Retry" capability to autonomously overcome ID Errors by learning a policy decoupled from $s_t^r$.
\item Include a library of learned ``Reset" skills, $\{\pi_{\text{reset}, j}\}$, that can be invoked to transform an OOD state $s_t^e \notin \mathcal{S}_{\text{task}}^e$ back into a valid state $s_{t'}^e \in \mathcal{S}_{\text{task}}^e$.
\end{enumerate}

\subsection{Enabling Robust "Retry" via Perturbation \& Bridging}
\label{sec:method_retry}

As mentioned above, a trajectory-monotonic nature of human demonstrations is the primary barrier to a VLA's innate ``retry" capability. While prior work has used segment stitching to generate new data~\cite{mandlekarmimicgen}, this approach alone fails to solve trajectory monotonicity. We address this by introducing a novel Perturbation \& Bridging augmentation pipeline that systematically decouples the robot pose ($s_t^r$) from the environment state ($s_t^e$), thereby building true robustness.  
Our augmentation process is as follows:

\vspace{1mm}
\noindent \textbf{Trajectory Composition} We first compose a new, high-level task trajectory from a small library of human-demonstrated subtask segments (e.g., ``grasp mug," ``place mug"). For a new, randomly sampled scene, we select and kinematically transform these segments to match the new object poses, stitching them together to form a baseline trajectory $\mathcal{T}_{\text{task}} = (a_0, a_1, \ldots, a_N)$, where $a_i$ is an action segment for a task step.

\vspace{1mm}
\noindent \textbf{Perturbation \& Bridging}: This baseline trajectory, which still suffers from monotonicity, is then fundamentally enhanced. We systematically inject ``perturbation-bridging" segments $d_i$ between the task segments, creating a new, robust trajectory: $\mathcal{T}_{\text{aug}} = (d_{\text{init}}, a_0, d_0, a_1, d_1, \ldots, a_N)$.

\vspace{1mm}
\noindent \textbf{Segment Structure}: Each segment $d_i$ consists of two phases:

\begin{itemize}
    \item Perturbation ($d_i^A$): A random motion sequence (e.g., random joint velocities or end-effector movements) that moves the robot arm to an arbitrary, out-of-distribution pose. This phase breaks the pose-state correlation.

    \item Bridging ($d_i^B$): A recovery action sequence that moves the robot from its perturbed pose (at the end of $d_i^A$) back to the valid starting pose required for the next task segment, $a_{i+1}$. Specifically, the bridging segment $d_i^B$ connects the robot's perturbed state (at the end of $d_i^A$) back to the valid starting pose required for the next task segment $a_{i+1}$. 
    
\end{itemize}

To generate these segments efficiently, we adopt the standard planner-free kinematic interpolation technique~(linear interpolation for position and SLERP for rotation)~\cite{mandlekarmimicgen}.
The benefit of generating a massive, diverse dataset of retry examples far outweighs the noise from rare, invalid trajectories, which is the key to learning the decoupled policy. 
The initial segment $d_{\text{init}}$ serves the same purpose, decoupling the very first action $a_0$ from a fixed starting pose.

\vspace{1mm}
\noindent \textbf{Training Data Generation}
Crucially, the perturbation actions $d_i^A$ are not used as training targets, as they are not meaningful actions. Instead, we form new training sequences by combining the bridging actions with the subsequent task actions. For each augmented trajectory, we add the following sequences to our training dataset:
\begin{itemize}
\item The original successful subsequences (e.g., $(a_i, a_{i+1}, \ldots)$).
\item The new bridging-to-task subsequences (e.g., $(d_i^B, a_{i+1}, a_{i+2}, \ldots)$).
\end{itemize}
By training on these $(d_i^B, a_{i+1}, \ldots)$ sequences, the VLA explicitly learns to execute the correct task action $a_{i+1}$ from a wide variety of initial robot poses (the end-states of $d_i^B$). This directly breaks the spurious pose-state coupling and builds in a generalizable, innate ``retry" capability.

\subsection{Learning "Reset" Skills via MLLM-driven Failure Mining}
\label{sec:method_reset}

The ``retry" capability only solves ID errors. It cannot handle OOD errors where the environment itself is in a catastrophic failure state. The ``success-centric" nature of demonstration datasets means VLAs are never exposed to such states or their corresponding recovery procedures.

We address this data scarcity with an efficient pipeline to mine, label, and learn object-centric ``reset" skills~(Fig.~\ref{fig:framework}). More specifically, the procedure is as follows:

\vspace{1mm}
\noindent \textbf{Failure Collection} We first execute our ``retry-capable" model (trained as in Sec.~\ref{sec:method_retry}) on its designated tasks over hundreds of episodes. We collect all execution videos and trajectory data, including both successes and failures.

\vspace{1mm}
\noindent \textbf{Offline MLLM Failure Analysis} We then use a Multi-modal LLM as an ``offline failure analyst" to process this large corpus of videos. For each failure video, the MLLM is prompted to provide a structured JSON output specifying:
\begin{itemize}
\item \textit{error\_type}: (``ID-Retryable" / ``OOD-Reset\_Required")
\item \textit{reset\_target}: (The name of the object that needs resetting, e.g., ``cup" or ``None")
\item \textit{error\_group}: (The objects that forms the error, e.g., the coffee pod stuck at the coffee pod holder)
\item \textit{failure\_timestamp}: (The time $t$ at which the OOD error occurred)
\end{itemize}

The \textit{error\_group} allows us to treat the interacting error objects (e.g., the ``stuck pod" and the ``holder") as a single ``semantic failure asset." 
When generating new reset demonstrations in the subsequent step, we can sample and place this entire \textit{error\_group} into a variety of new scene configurations, while preserving the relative poses that define the failure state $s_t$. This ensures our VLA learns to recover from the semantic nature of the error (the ``stuck" state) regardless of the failure's position in the scene, which is essential for training generalizable reset skills.

\vspace{1mm}\noindent \textbf{Bootstrapping the Reset Dataset} This automated MLLM analysis bootstraps our reset skill collection by providing a high-quality corpus of OOD failure configurations. For each failure identified as OOD-Reset\_Required, the failure\_timestamp $t$ allows us to extract the precise, high-fidelity environment state $s_t$ (e.g., the MuJoCo state) from our logs.

This mined dataset consists of (error state, skill label) pairs. The error state $s_t$ captures the full physical configuration of the \textit{error\_group} (e.g., $s_{\text{pod\_stuck}}$ describes the pod's exact pose relative to the machine holder). The skill label is derived from the \textit{reset\_target} (e.g., ``reset the pod"). This corpus of ($s_t$, $\mathcal{I}_{\text{reset}}$) pairs serves as the starting configurations for the next step: for human demonstration collection.

\vspace{1mm}
\noindent \textbf{Demonstrations and Augmentation}
For each identified object-centric reset skill (e.g., ``reset the cup"), we collect a small set (e.g., 10-20) of human demonstrations, $\mathcal{D}_{\text{reset\_human}}$. Each demonstration begins from a corresponding mined error state $s_t$, and the operator provides a recovery trajectory $\tau_{\text{reset}}$ to restore the \textit{reset\_target} to a valid state.

This small, curated dataset is then massively expanded. To generate a new, robust demonstration, our pipeline first preserves the semantic integrity of the failure by placing the entire \textit{error\_group} (maintaining its internal relative poses) into a new scene. Then, it applies our same Perturbation \& Bridging augmentation to this $\tau_{\text{reset}}$ trajectory.

This process generates a large-scale dataset $\mathcal{D}_{\text{reset\_aug}}$ that makes our ``reset" skills robust not only to the robot's starting pose but also to the contextual layout of the failure itself.

\subsection{Unified Training and Closed-Loop Inference} \label{sec:method_system}

Instead of training a single, monolithic model prone to task interference, we adopt a modular and scalable expert policy library approach. We leverage Low-Rank Adaptation~(LoRA)~\cite{hu2022lora} to efficiently train a set of specialized policy adapters on top of a shared, pre-trained VLA backbone. This design allows each policy to achieve high performance on its specific task while enabling straightforward system-level scaling.

\vspace{1mm}
\noindent \textbf{Training an Expert Policy Library}
Our unified training dataset, comprising $\mathcal{D}_{\text{task\_aug}}$ (original and ``retry-augmented" task data) and $\mathcal{D}_{\text{reset\_aug}}$ (all augmented ``reset" skill data), is used to train this library. Specifically:
\begin{itemize}
\item A main task adapter, $\pi_{\text{LoRA}}^{\text{task}}$, is trained on $\mathcal{D}_{\text{task\_aug}}$ using the main task instruction (e.g., $\mathcal{I}_{\text{task}}$ = ``make coffee").
\item A distinct reset skill adapter, $\pi_{\text{LoRA}}^{\text{reset}, j}$, is trained for each object-centric reset skill $j$. For example, the ``reset cup" adapter is trained exclusively on its corresponding reset demonstrations, using the prompt $\mathcal{I}_{\text{reset}}$ = ``reset the cup."
\end{itemize}
This modular approach prevents the conflicting gradients that can arise from training disparate skills within a single set of weights. Furthermore, it provides a clear path for extensibility, as new recovery skills can be added to the system by simply training new LoRA adapters, without disturbing existing policies.

\vspace{1mm}
\noindent \textbf{Closed-Loop Orchestration}
In deployment, the system operates as a closed loop orchestrated by the online MLLM monitor~(Fig.~\ref{fig:framework}).
\begin{itemize}
\item The system initializes by loading the main task adapter, $\pi_{\text{LoRA}}^{\text{task}}$, and begins execution with the prompt $\mathcal{I}_{\text{task}}$.
\item The MLLM monitor observes the VLA's execution.
\item \textbf{If an ID Error occurs}: No intervention is needed. The VLA's innate ``retry" robustness, learned from our perturbation augmentation, allows it to autonomously recover and continue the task.
\item \textbf{If an OOD Error occurs}: The MLLM intervenes. It identifies the required reset skill (e.g., ``reset the cup") and directs the control system to swap the active LoRA adapter to the corresponding $\pi_{\text{LoRA}}^{\text{reset}, j}$.
\item The VLA, now equipped with its ``reset" expert policy, executes the recovery skill.
\item Once the MLLM confirms the environment is reset to a valid state, it directs the system to reload the main task adapter $\pi_{\text{LoRA}}^{\text{task}}$. The VLA then seamlessly resumes the main task.
\end{itemize}
This two-level design, where a high-level MLLM directs a library of specialized VLA adapters, results in a system that is robust, high-performing, and easy to scale.

\begin{table*}[t]
\caption{Comparison of experimental results across 9 manipulation tasks in RoboMimic Simulation. The `D' suffix in the task names denotes the range of object randomization in scene initialization. The results demonstrate that our failure-aware VLA framework for robust robotic manipulation achieves better performance than all previous methods as well as the strong VLA baselines. Phoenix-Human is gray as it involves human instruction as the self-reflection subgoal.}
\label{tab:main_results}
\centering
\resizebox{\textwidth}{!}{
\begin{tabular}{ccccccccccc}
\toprule
Methods & Coffee\_D0 & Coffee\_D1  & Stack\_D0 & Stack\_D1 & StackThree\_D0 & StackThree\_D1 & Threading\_D0 & \makecell{ThreePiece\\Assembly\_D0} & \makecell{ThreePiece\\Assembly\_D1} & Mean\\ \midrule

OpenVLA~\cite{kim2025openvla} & 42\% & 18\% & 84\% & 86\% & 36\% & 20\%  & 20\%  & 28\% & 8\% & 38.0\%    \\

Task-conditioned & 66\% & 24\% & 88\% & 68\% & 30\% & 6\% & 74\% & 20\% & 0\% & 41.8\%    \\
Subgoal-conditioned & 76\% & 26\% & 88\% & 74\% & 24\% & 6\% & \textbf{78\%} & 20\% & 2\% & 43.8\%  \\
Motion-conditioned & 68\% & 32\% & 92\% & 84\% & 38\% & 16\% & 58\% & 30\% & 4\% & 46.9\%  \\
\midrule

Subgoal Self-reflection & 80\% & 32\% & 88\% & 78\% & 32\% & 6\% & 80\% & 34\% & 2\% & 48.0\%    \\

Phoenix~\cite{xia2025phoenix} &   {94\%} & {48\%} & {96\%} &{86\%} & {50\%} & {20\%} & 68\% & {52\%} & 6\% & {57.8\%} \\

{\color{gray} Phoenix-Human} & {\color{gray} 100\%} & {\color{gray} 100\%} & {\color{gray} 100\%} & {\color{gray} 90\%} & {\color{gray} 70\%} & {\color{gray} 40\%} & {\color{gray} 100\%} & {\color{gray} 70\%} & {\color{gray} 40\%} & {\color{gray} 78.9\%} \\

\midrule

$\pi_{0.5}$~\cite{intelligence2025pi_} &   {82\%} & {56\%} & \textbf{100\%} & 92\% & 90\% & 84\% & 42\% & {58\%} & 46\% & 72.2\% \\

Ours &   \textbf{96\%} & \textbf{78\%} & \textbf{100\%} &\textbf{100\%} & \textbf{100\%} & \textbf{90\%} & 72\% & \textbf{62\%} & \textbf{58\%} & \textbf{84.0\%} \\

\bottomrule
\end{tabular}}
\end{table*}

\section{Experiment}

\label{sec:experiment}

In this work, we conduct experiments on 9 contact-rich manipulation tasks in RoboMimic~\cite{mandlekar2022matters} following previous studies~\cite{xia2025phoenix}.
The detailed description for each task can be found in \cite{mandlekar2022matters} and in our supplementary materials.

\vspace{1mm}
\noindent \textbf{Implementation Details}
We generated 500 augmented demonstrations per task from 10 initial human-collected demos using Mimicgen~\cite{mandlekarmimicgen}. Our ``retry" augmentation injects perturbation \& bridging sequences (max $45^{\circ}$ rotation, $0.5$m translation) at the trajectory start and between subtasks. For ``reset" skills, we used Gemini-2.5-Pro to analyze failure videos. We collected 20 human demos for each object-centric reset skill and augmented them to 500 using the same perturbation method. We fine-tuned the language model and action expert of $\pi_{0.5}$~\cite{intelligence2025pi_} using LoRA~\cite{hu2022lora}, training with the Adam optimizer at a constant learning rate of $2.5 \times 10^{-4}$ without decay. All results are averaged over 50 evaluation trials per task following \cite{xia2025phoenix}. Further details are available in our supplementary materials.

\vspace{1mm}
\noindent \textbf{Baselines} Following previous studies~\cite{xia2025phoenix}, we compare our method to the following baselines in three aspects, powerful open-source VLA baselines, conditional diffusion policy, and self-reflection policy. More details are available in the supplementary materials.
\begin{itemize}
    \item \textbf{OpenVLA~\cite{kim2025openvla}.} OpenVLA is fine-tuned to provide baseline performance for multi-task experiments.
    \item \textbf{Task-conditioned policy~\cite{xia2025phoenix}.} The diffusion policy is trained with the task description as the condition.
    \item \textbf{Subgoal-conditioned policy~\cite{xia2025phoenix}.} A LLaVAv1.5 is fine-tuned to predict subgoals at 5Hz, which are utilized as the condition for diffusion policy.
    \item \textbf{Motion-conditioned policy~\cite{xia2025phoenix}.} A LLaVA-v1.5 is fine-tuned as the subgoal self-reflection model and applied to the subgoal-condition policy.
    \item \textbf{Phoenix~\cite{xia2025phoenix}.} A motion-based self-reflection framework for action correction.
    \item \textbf{Phoenix-Human~\cite{xia2025phoenix}.} This method provides an upper bound on the performance of self-reflection methods by manually correcting the wrong motion instructions.
    \item \textbf{$\pi_{0.5}$~\cite{intelligence2025pi_}.} The successor to the $\pi_0$ VLA model~\cite{black2024pi_0}, which demonstrates improved generalization by training on diverse, multi-environment datasets. It serves as the VLA backbone for our method. 
\end{itemize}

\vspace{2mm}
\noindent \textbf{Comparison Results} 
The overall results are summarized in Table~\ref{tab:main_results}. Our method achieves state-of-the-art performance on 8 out of 9 tasks. On the remaining task, Threading\_D0, although we do not obtain the best result, our approach still (1) substantially boosts the performance of the VLA backbone $\pi_{0.5}$ and (2) outperforms Phoenix~\cite{xia2025phoenix}, the previous leading method for error recovery.

Our method achieves an average success rate of 84.0\%, significantly surpassing Phoenix's 57.8\%~\cite{xia2025phoenix}. While part of this gain comes from the strong capability of the $\pi_{0.5}$ backbone, our approach still provides an additional 11.8\% improvement over $\pi_{0.5}$ alone. More notably, our method even outperforms Phoenix-Human, demonstrating the comprehensive advantage of our framework over prior self-reflection approaches—even when compared to a baseline supplied with correct human guidance.

For the Stack task, the reset setting is not applicable because the task involves stacking two cubes of different colors, and there are almost no non-retryable errors during execution. In this case, our method still achieves comparable performance, even when multiple baselines (Phoenix-Human, $\pi_{0.5}$) reach a 100\% success rate.

The `D' suffix in the task names (e.g., D0, D1) denotes the range of object randomization during scene initialization and therefore reflects task difficulty. For example, Coffee\_D0 contains less variation than Coffee\_D1 and is thus easier. Across most tasks, the performance improvement provided by our method on the D1 versions is more pronounced than on the D0 versions. We attribute this to the decoupling effect of our perturbation-and-bridging strategy, which enhances the VLA model's robustness to environmental variations.

\begin{figure}
\centering
    \includegraphics[width=\linewidth]{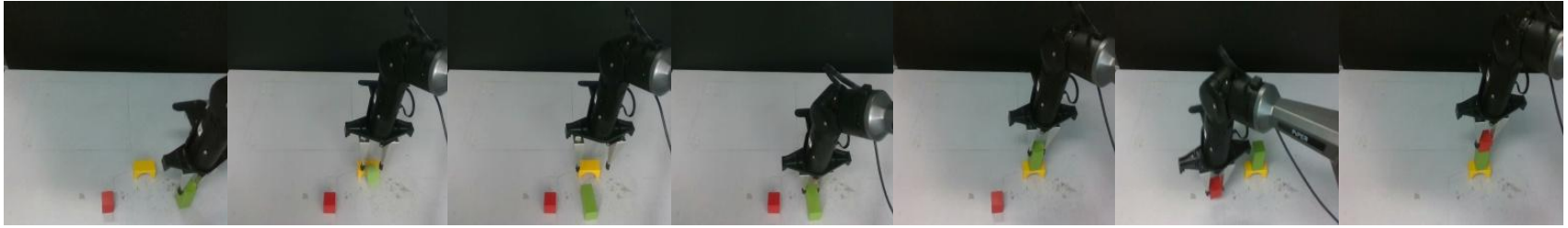} \\
    
    \vspace{-0.2mm} 
    
    \includegraphics[width=\linewidth]{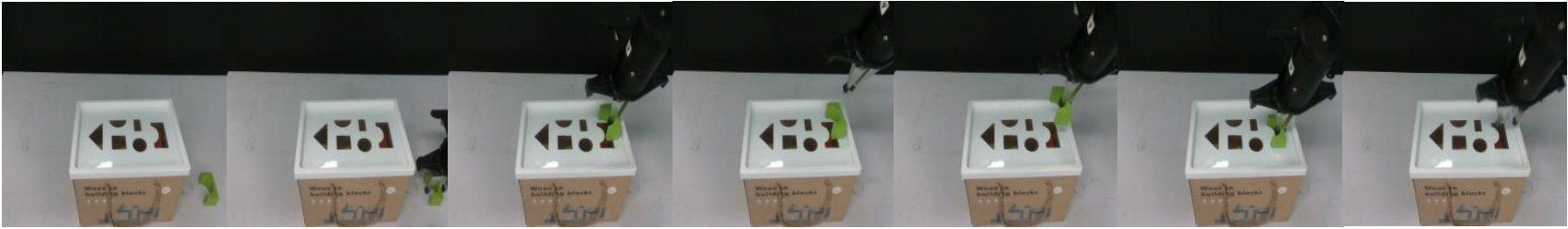}
\caption{Real-world experimental setup and execution trajectories. \textbf{Top:} The ``Stack Three Blocks'' task. \textbf{Bottom:} The ``Insert U-shaped Block'' task. }
\label{fig:real_world}
\end{figure}

\noindent \textbf{Real-world Validation} 
To verify FLARE's effectiveness and address concerns about privileged simulation states, we conducted real-world experiments on a Piper arm with RealSense D435i~(top/wrist views) across two challenging tasks: Stack Three Blocks (long-horizon) and Insert U-shaped Block (contact-rich). The experimental results are presented in Table~\ref{tab:real_world} and Fig.~\ref{fig:real_world}.
We collected only 10 human demos and augmented them to 50 for each task/reset skill. To bypass the need for ground-truth simulator states, we used Any6D~\cite{lee2025any6d} for object pose estimation to facilitate data augmentation. This proves that our method can be deployed using only visual sensors without privileged coordinate logs. For failure reconstruction, strict coordinate matching is unnecessary; FLARE relies on approximate failure reconstruction. Collecting roughly similar failure states (e.g., a toppled cup) without precise spatial alignment is sufficient, as our perturbation-based augmentation ensures generalization to diverse real-world poses using only visual feedback.

\begin{table}[t]
\caption{Success rates of real-world manipulation tasks. Our method outperforms the baseline in two challenging settings.}
\label{tab:real_world}
\centering
\resizebox{0.9\linewidth}{!}{
\begin{tabular}{lcc}
\toprule
\textbf{Real-World Task~(40 trials)} & \textbf{$\pi_{0.5}$ (Baseline)} & \textbf{Ours (FLARE)} \\
\midrule
Stack Three Blocks & 62.5\% & \textbf{75.0\%} \\
Insert U-shaped Block & 45.0\% & \textbf{55.0\%} \\
\bottomrule
\end{tabular}
}
\end{table}

\section{Ablation and Analysis}

\subsection{Analysis of Perturbation \& Bridging}

\begin{figure}
    \centering
    \includegraphics[width=\linewidth]{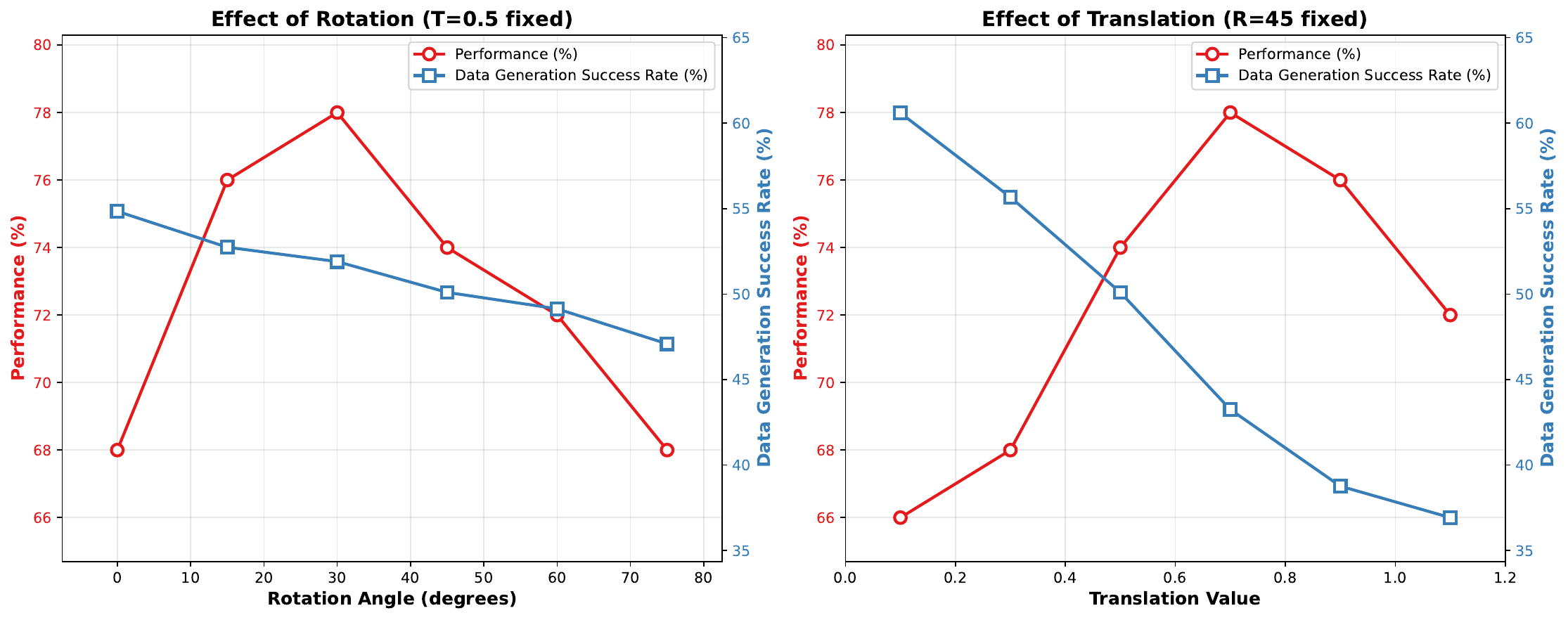}
    \caption{The sensitivity of two hyper-parameters, rotation $r$ and translation $t$ for perturbation \& bridging. Our method achieves best performance when $r=30^\circ$ and $t = 0.7$, respectively.}
    \label{fig:retry_analysis}
\end{figure}

\noindent \textbf{The Sensitivity of Hyper-Parameters}
In this section, we investigate two key hyperparameters in our perturbation \& bridging augmentation strategy: the rotation $r$ and the translation $t$. The rotation $r$ controls the rotation angle applied to the robot hand during perturbation, while the translation $t$ determines the positional shift. We generate retry demonstrations under different combinations of $r$ and $t$, and fine-tune the VLA model using these demonstrations. All experiments are conducted on Coffee\_D1, and the results are presented in Fig.~\ref{fig:retry_analysis}.

As shown in Fig.~\ref{fig:retry_analysis}, our method consistently outperforms all baselines, including Phoenix and $\pi_{0.5}$. The best performance is achieved when $r=30^\circ$ and $t=0.7$ in the two settings, respectively. We also report the corresponding demonstration-generation success rate for each setting, which reflects the efficiency of generating valid retry demonstrations. As illustrated, larger rotations and translations produce demonstrations with higher variance and therefore yield higher task success rates, but at the cost of reduced generation efficiency.
Excessively large perturbations eventually degrade performance. A plausible explanation is that the training set contains a limited number of demonstrations, and overly high variance may hinder the model's ability to generalize effectively.

\begin{figure}[t]
    \centering

    \begin{subfigure}{0.25\linewidth} 
        \centering
        \includegraphics[width=\linewidth]{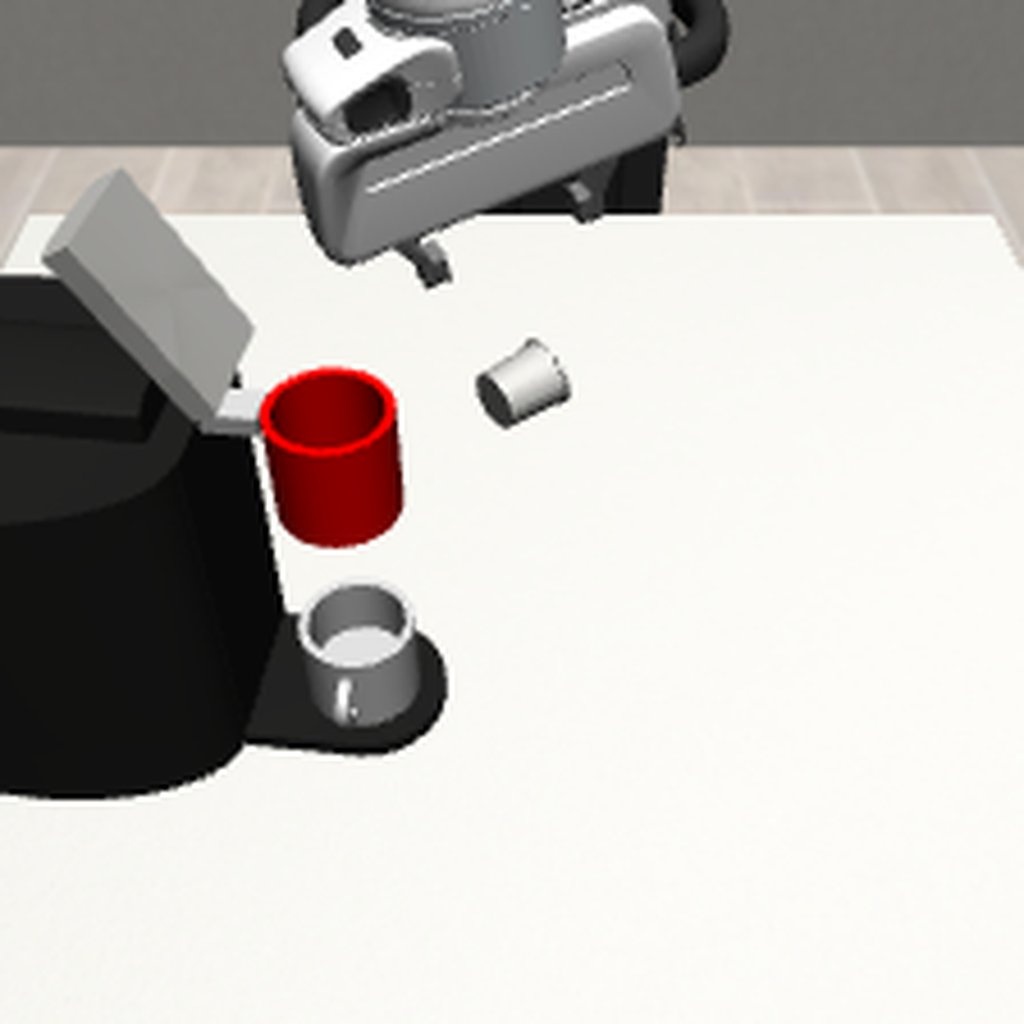}
        \caption{}
        \label{fig:sub1}
    \end{subfigure}
    \begin{subfigure}{0.25\linewidth} 
        \centering
        \includegraphics[width=\linewidth]{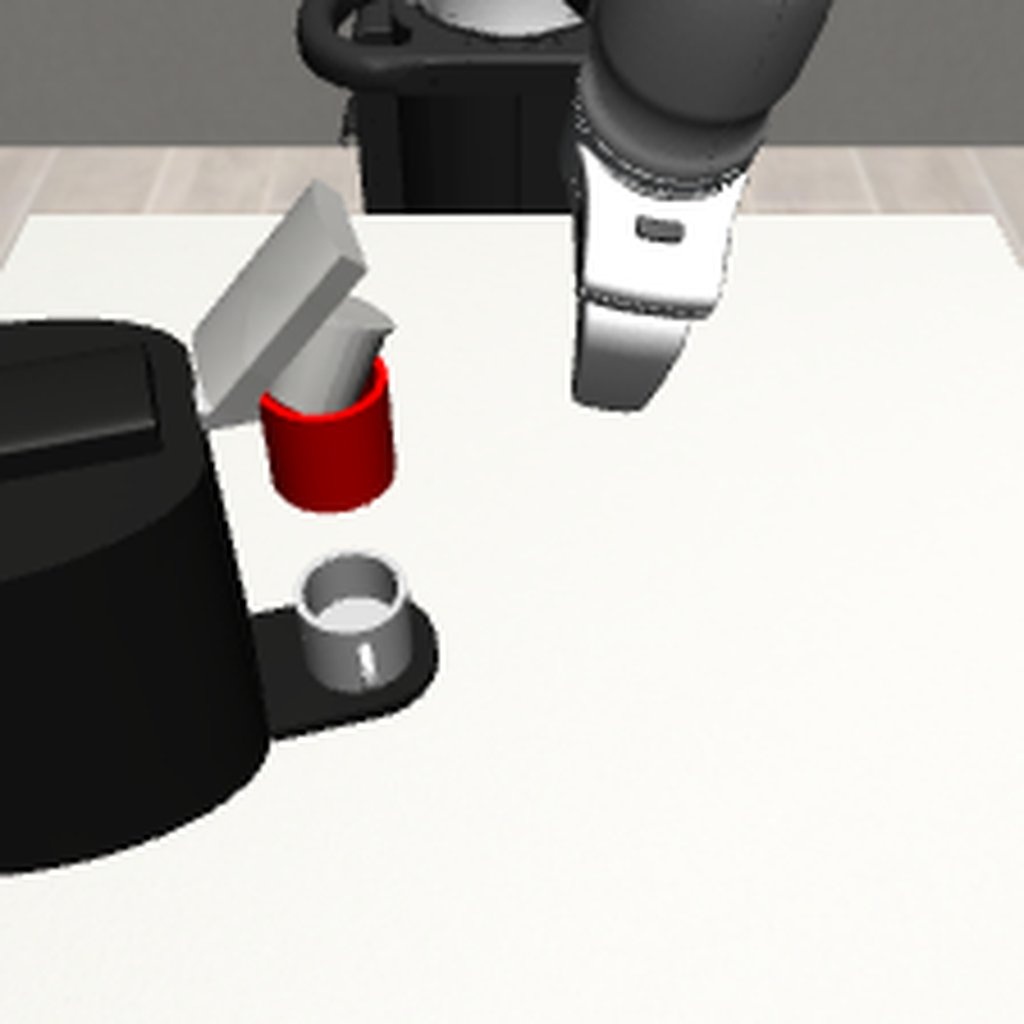}
        \caption{}
        \label{fig:sub2}
    \end{subfigure}%
    \begin{subfigure}{0.25\linewidth} 
        \centering
        \includegraphics[width=\linewidth]{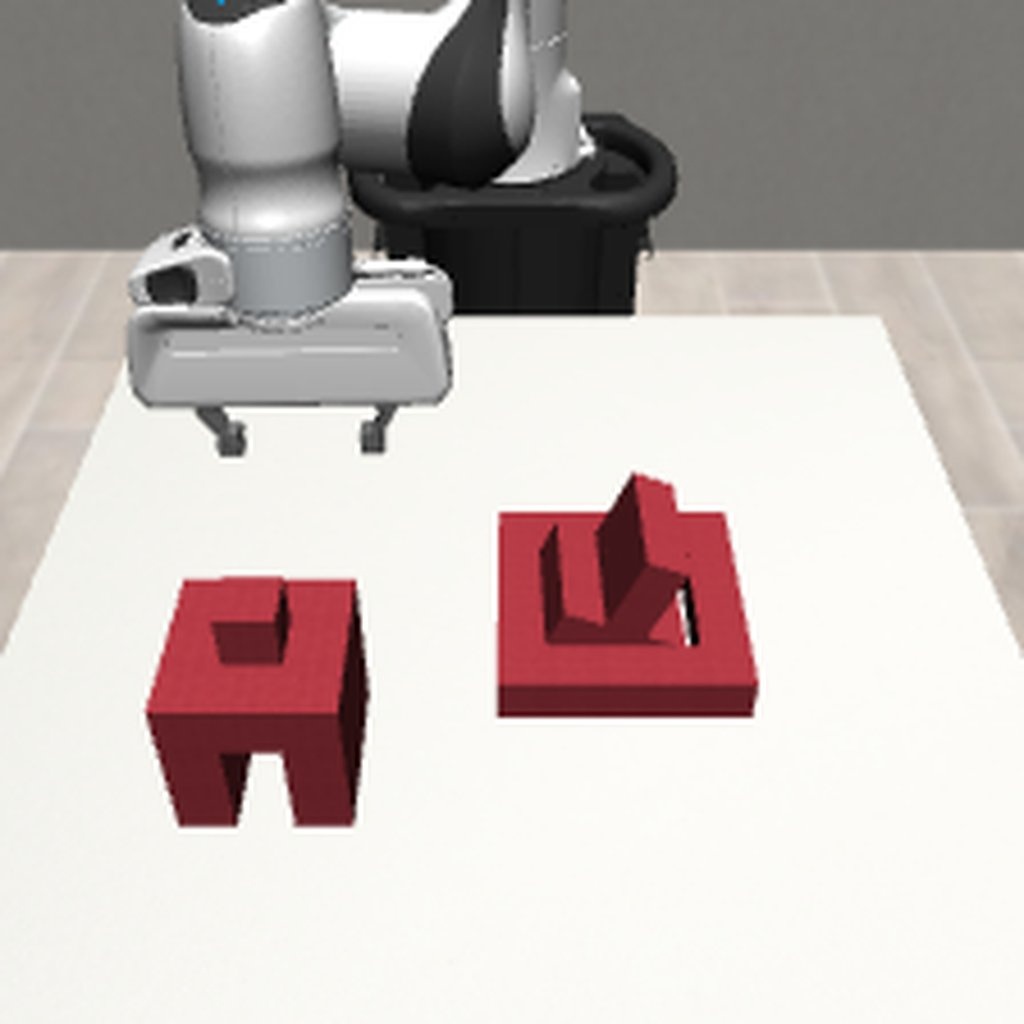}
        \caption{}
        \label{fig:sub3}
    \end{subfigure}
    \begin{subfigure}{0.25\linewidth} 
        \centering
        \includegraphics[width=\linewidth]{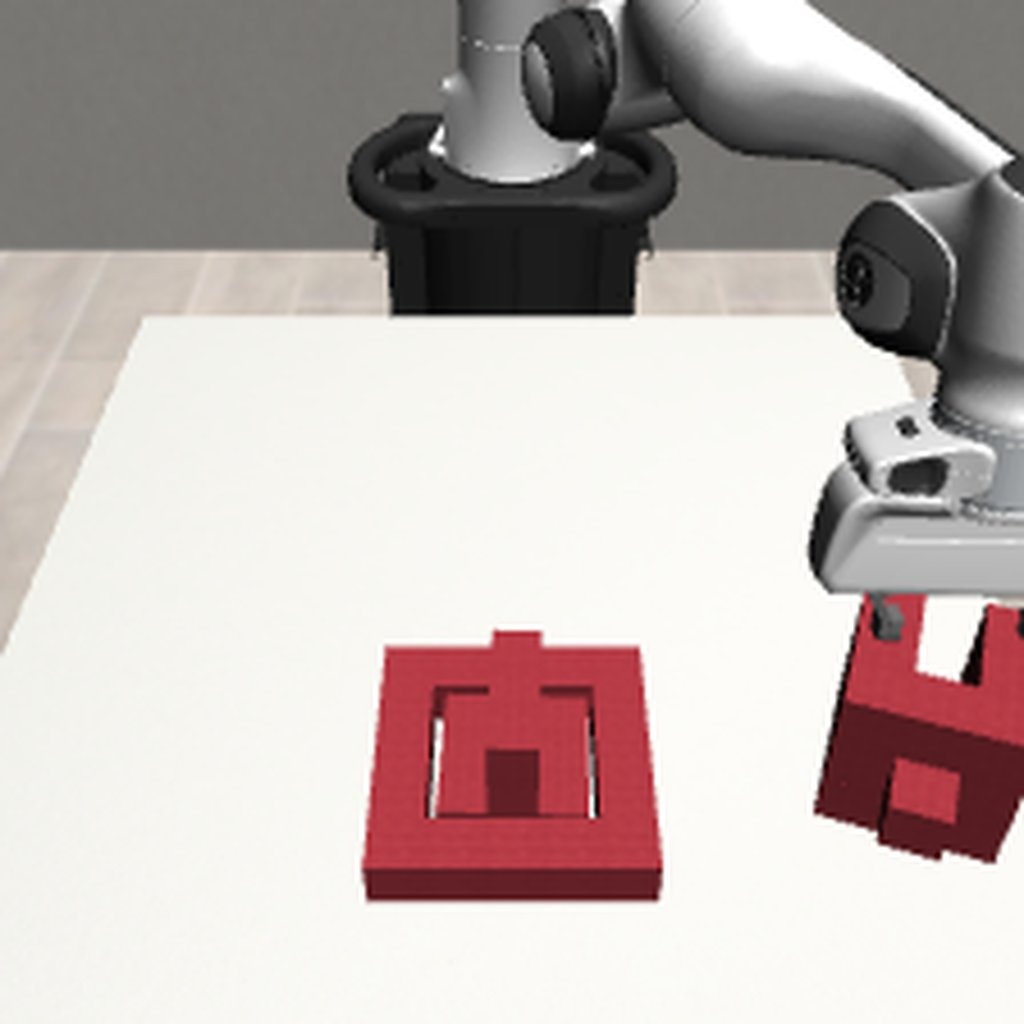}
        \caption{}
        \label{fig:sub4}
    \end{subfigure}

    \caption{The failure case for Coffee and ThreePieceAssembly respectively. (a) Reset coffee pod; (b) Reset coffee machine lid; (c) Reset the T-shaped block; (d) Reset the U-shaped block.}
    \label{fig:failure_case}
\end{figure}

\subsection{Ablations and Analysis for Reset skills learning}

\noindent \textbf{Ablation of Reset and Human Instruction}
In this section, we first investigate the importance of reset-skill learning and its application within our framework.
Following previous studies~\cite{xia2025phoenix}, we adopt a multimodal LLM to generate task prompt that decides which reset skill should be executed based on the current environment state. To assess the necessity of this component, we ablate the reset skill entirely and also evaluate a variant of our framework that replaces the multimodal LLM with human-provided instructions, serving as an upper bound.

The results are reported in Table~\ref{tab:ablation}, covering four settings—``Ours'', ``Ours w/o Reset'', ``Ours Reset-Only'' and ``Ours-Oracle''—on the Coffee and ThreePieceAssembly tasks. The comparison illustrates the effectiveness of reset-skill learning and deployment: removing reset skills reduces the average success rate by 3.5\%. The human-instruction variant (Ours-Oracle) achieves an additional 7\% improvement, suggesting that stronger multimodal LLMs could further enhance task execution monitoring.

\begin{table}[t]
\caption{The performance comparison for our method and two variants, Ours w/o Reset and Ours-Oracle. Ours w/o Reset only applies the perturbation \& bridging strategy for VLA model training, and Ours-Oracle replaces the generated instruction for object reset with human feedback. The experimental results demonstrate the effectiveness of our reset skill learning and error correction.}
\label{tab:ablation}
\centering
\resizebox{\columnwidth}{!}{
\begin{tabular}{ccccc}
\toprule
Methods & Coffee\_D0 & Coffee\_D1 & \makecell{ThreePiece\\Assembly\_D0} & \makecell{ThreePiece\\Assembly\_D1} \\ \midrule

Ours w/o Reset &   92\% & 74\% & 60\% & 54\%  \\

Ours Reset-Only &   88\% & 64\% & 60\% & 50\%  \\

Ours &   96\% & 78\% & 62\% & 58\% \\

Ours-Oracle &   100\% & {90\%} & 68\% & 64\%  \\

\bottomrule
\end{tabular}
}
\end{table}

\vspace{2mm}
\noindent \textbf{Reset Skill Learning Analysis} 
We further analyze reset-skill learning by evaluating both task success rate and demonstration-generation efficiency. Specifically, we select two tasks—Coffee and ThreePieceAssembly—and examine the two reset objects required for each. For Coffee, the reset objects are the coffee machine lid (Object 1) and the coffee pod (Object 2). For ThreePieceAssembly, the reset objects correspond to the T-shaped block (Object 1) and the U-shaped block (Object 2).
We construct two dedicated environments for reset evaluation, redefine the success conditions based on the reset target, and apply domain randomization during scene initialization. Fig.~\ref{fig:failure_case} visualizes the four reset tasks. Note that in Fig.~\ref{fig:sub2}, the robot must reset the coffee machine lid before adjusting the coffee pod pose; thus, the effective reset object is the lid.

The results are shown in Table~\ref{tab:reset_success_rate}. For Coffee, the success rate for resetting the machine lid reaches 84\%, substantially higher than the coffee pod reset. A similar pattern appears in the ThreePieceAssembly tasks, where the T-shaped block reset is significantly easier than resetting the U-shaped block. Resetting the coffee pod, for example, often requires grasping a toppled pod, adjusting its pose, and placing it upright—an operation that is challenging for the robot hand used in our experiments. These findings suggest the need for more advanced manipulation capabilities for pose adjustment in VLA-based error correction, such as dexterous in-hand pose refinement.

We also report demonstration-generation efficiency in Table~\ref{tab:reset_success_rate}. As expected, efficiency is lower for more difficult reset tasks, matching the success-rate trend. The reduced success rate indicates room for further augmentation and model training, which could potentially enhance error-correction performance.

\vspace{2mm}
\noindent \textbf{Failure Case Identification with Multi-modal LLM}
To evaluate the performance of failure-case identification, we manually label 50 execution videos for the Coffee and ThreePieceAssembly and measure accuracy across three components: retry/reset classification, reset-object identification, and timestamp identification. As described in Sec.~\ref{sec:experiment}, we employ Gemini-2.5-Pro as the multimodal LLM, and the exact prompts are provided in the supplementary materials.

The retry/reset classification accuracy reaches 88\% and 96\% for the two tasks, demonstrating the strong video reasoning capabilities of state-of-the-art multimodal LLMs. Reset-object identification is a two-way classification problem for each task, achieving accuracies of 88\% and 78\%.
For timestamp identification, we compute accuracy by checking whether the identified frame corresponds to the correct reset object.

\begin{table}[t]
\caption{The reset skill success rate and demonstration generation efficiency on two manipulation tasks. The Reset Object 1 is coffee machine lid/T-shaped block, and the reset object 2 is coffee pod/U-shaped block for Coffee/ThreePieceAssembly. Our method successfully generates demonstrations and trains useful VLA models that can reset the objects when errors occur.}
\label{tab:reset_success_rate}
\centering
\resizebox{\columnwidth}{!}{
\begin{tabular}{l cccc}
\toprule
\multicolumn{1}{c}{} & \multicolumn{2}{c}{Reset Skill Success Rate} & \multicolumn{2}{c}{Generation Efficiency} \\
\cmidrule(lr){2-3} \cmidrule(lr){4-5} 

Task & Coffee & \makecell{ThreePiece\\Assembly} & Coffee & \makecell{ThreePiece\\Assembly} \\ 
\midrule

Reset Object 1 & 84\% & 88\% & 83.7\% & 48.6\% \\
Reset Object 2 & 24\% & {20\%} & 11.6\% & 5.9\% \\

\bottomrule
\end{tabular}
}
\end{table}

\begin{table}[t]
\caption{The performance of failure case indentification of Gemini-2.5-Pro on three tasks: Reset/Retry classification, Reset object identification and Timestamp identification for Coffee and ThreePieceAssembly. The effective failure analysis provides high-quality reset dataset for further demonstration collection.}
\label{tab:failure_case_identification}
\centering
\resizebox{\columnwidth}{!}{
\begin{tabular}{cccc}
\toprule
Task & Reset/Retry & Reset Object & Timestamp \\ \midrule
Coffee &  88\% & 88\% & 78\%  \\
ThreePiece Assembly &   96\% & 78\% & 66\% \\
\bottomrule
\end{tabular}
}
\end{table}

\section{Conclusion}
We presented FLARE, a failure-aware framework that endows VLA agents with robust autonomy through a dual Retry/Reset paradigm. We identified that the brittleness of current VLAs stems from trajectory-monotonic training data and success-biased distributions. To overcome this, we proposed a perturbation-based augmentation to facilitate innate ID retry capabilities and an MLLM-guided pipeline to acquire OOD reset skills. Our approach achieves an 84.0\% average success rate across diverse manipulation tasks, surpassing existing self-reflection methods and VLA backbones. While current hardware limits the correction of highly complex object poses, our findings confirm that treating failure recovery as a distinct, learned capability is essential for open-world robotic deployment. Future research will focus on scaling reset skill libraries and enhancing dexterous recovery maneuvers.

\subsection*{Acknowledgments}
This work is sponsored by CIE-Tencent Robotics X Rhino-Bird Focused Research Program and is also supported by the National Natural Science Foundation of China (NO.~62322608).

{
    \small
    \bibliographystyle{ieeenat_fullname}
    \bibliography{main}
}

\clearpage
\setcounter{page}{1}
\maketitlesupplementary

\setcounter{section}{0}

\section{Implementation Details}

\noindent \textbf{Data Generation and Augmentation.} For our primary task data, we use MimicGen~\cite{mandlekarmimicgen} with its default parameters. For each task, we begin with a set of 10 human-collected demonstrations that are then expanded to 500 augmented demonstrations. Our perturbation and bridging process, critical for enabling the ``retry" capability, is defined by two parameters: a maximum perturbation rotation of $45^{\circ}$ and a maximum translation of $0.5$ meters. Following the MimicGen procedure, we segment each task trajectory into subtasks. Our perturbation \& bridging sequences are then injected at the beginning of the full trajectory and at the junctions between these subtasks.

\vspace{2mm}
\noindent \textbf{Reset Skill Collection.} Our reset skills are designed to be object-centric. The total number of reset skills for a given task is determined by the number of manipulable objects involved. We explicitly scope out unrecoverable situations, such as when an object is too large for the gripper to re-orient (e.g., a toppled coffee machine) or when an object falls outside the robot's reachable workspace (e.g., onto the floor).To mine failure states for reset demonstration, we use Gemini-2.5-Pro to analyze task execution videos, with a sampling temperature set to $0.7$. For each identified object-centric reset skill, we collect 20 human demonstrations. This initial dataset is then augmented to 500 demonstrations using the same perturbation \& bridging technique applied to the task data, ensuring our reset skills are also robust to varying robot poses.

\vspace{2mm}
\noindent \textbf{Training Details.} We follow the official training practices of $\pi_{0.5}$~\cite{intelligence2025pi_}. Specifically, we apply LoRA~\cite{hu2022lora} fine-tuning to the language model and the action expert modules, while all other model parameters remain frozen. We use the Adam optimizer with a constant learning rate of $2.5 \times 10^{-4}$ and do not employ a cosine learning rate decay schedule.

\vspace{2mm}
\noindent \textbf{Evaluation Protocol.} For each task, we conducted 50 evaluation trials and reported the average success rate, following established evaluation protocols in prior work~\cite{xia2025phoenix}.

\section{Manipulation Tasks}

\begin{figure}[t]
    \centering

    \begin{subfigure}{0.32\linewidth}
        \centering
        \includegraphics[width=\linewidth]{fig/demo_16_failure_timestep_7.jpg}
        \caption{}
        \label{fig:sub1}
    \end{subfigure}%
    \hfill
    \begin{subfigure}{0.32\linewidth}
        \centering
        \includegraphics[width=\linewidth]{fig/demo_42_failure_timestep_15.jpg}
        \caption{}
        \label{fig:sub2}
    \end{subfigure}%
    \hfill
    \begin{subfigure}{0.32\linewidth}
        \centering
        \includegraphics[width=\linewidth]{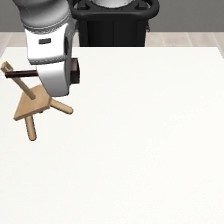}
        \caption{}
        \label{fig:sub3}
    \end{subfigure}

    \vspace{0.5\baselineskip}
    
    \begin{subfigure}{0.32\linewidth}
        \centering
        \includegraphics[width=\linewidth]{fig/demo_46_failure_timestep_8.jpg}
        \caption{}
        \label{fig:sub4}
    \end{subfigure}%
    \hfill
    \begin{subfigure}{0.32\linewidth}
        \centering
        \includegraphics[width=\linewidth]{fig/demo_65_failure_timestep_17.jpg}
        \caption{}
        \label{fig:sub5}
    \end{subfigure}%
    \hfill
    \begin{subfigure}{0.32\linewidth}
        \centering
        \includegraphics[width=\linewidth]{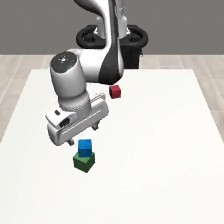}
        \caption{}
        \label{fig:sub6}
    \end{subfigure}

    \caption{Failure cases for all tasks (stack task excluded as reset is not applicable). (a) Reset coffee pod; (b) Reset coffee machine lid; (c) Reset the needle-hole block;  (d) Reset the T-shaped block; (e) Reset the U-shaped block; (f) Reset the blue block.}
    \label{fig:all_failure_case}
\end{figure}

In this section, we briefly describe the 9 manipulation tasks in our evaluation, and the reset task object for demonstration collection in our experiments. Fig.~\ref{fig:all_failure_case} shows the visualization of all these tasks.

\subsection{Visualization of Task Execution}
\begin{figure*}[!t]
    \centering
    \includegraphics[width=0.9\linewidth]{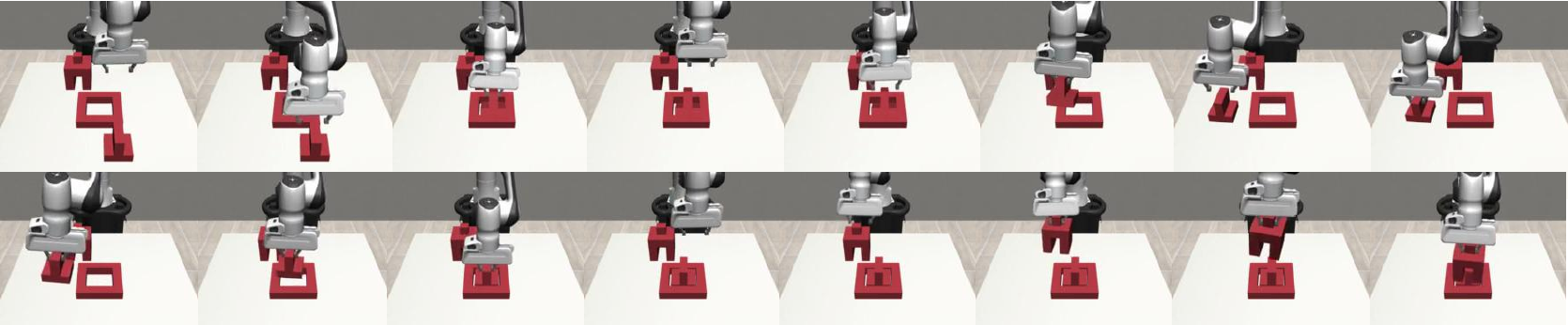}
    \caption{Visualization of a successful task execution for ThreePieceAssembly. Frame 1-4: The robot operates the regular task to assembly the three pieces by first pick-and-place the T-shaped block. Frame 5-8: The T-shaped block is not place appropriately and robot tries to reset it. Frame 9-16: The robot executes the regular task again and succeeds.}
    \label{fig:visualization}
\end{figure*}

We visualize a successful execution of the ThreePieceAssembly task and the corresponding reset instructions in Fig.~\ref{fig:visualization}. In this example, the robot first attempts to perform the regular task by picking up the T-shaped block and placing it into the square block. However, the T-shaped block is not placed correctly, triggering a reset. The robot then picks up the T-shaped block and places it aside to reset its pose. After completing the reset, the robot retries the task and successfully completes the assembly.

\section{Prompt for Video Analysis and Failure Detection}

In this section, we provide the prompts for both video analysis and failure detection during closed-loop inference in Table~\ref{tab:prompt_for_video_analysis} and Table~\ref{tab:prompt_for_closed_loop_inference}. 

\begin{table*}[t] 
    \centering
\caption{The main prompt used in our experiment for video analysis.}
\label{tab:prompt_for_video_analysis}
\begin{tcolorbox}[title=Video Analysis Query, width=\textwidth]
  \footnotesize
  You are an expert in robotic task failure analysis.  
Your task is to analyze a video recording of a robot performing a task that ultimately **failed**, and determine the **failure reason**, **recovery strategy**, **failure time point**, and **error group**.

---

\#\# Background
- The video shows the complete execution process of a robotic task that ended in failure.  
- You need to carefully inspect the entire sequence and identify the **key moment** and **cause** of failure.

---

\#\# Resettable Objects
\{reset\_objects\}

---

\#\# Concept of *Error Group*
An **Error Group** refers to the set of objects that jointly contribute to an erroneous state. Examples:

1. **Single-object error:**  
   The coffee capsule is overturned.  
   → Error Group = [``coffee capsule"]

2. **Multi-object error:**  
   The coffee capsule is jammed in the coffee machine’s insertion slot.  
   → Error Group = [``coffee capsule", ``coffee machine"]

When two or more objects form an undesirable **geometric or dynamic relationship** (e.g., a part stuck in a slot, an object wedged in a mechanism, or a collision with the machine workspace), the error involves multiple objects.  
To reproduce or reset such an error, **all objects in the group must be transformed together** (i.e., their relative poses must remain consistent).  
You cannot reset one object independently of the others.  

$>$ Note: The reset\_object must be included in the error\_group list.

---

\#\# Reset Selection Rule (Very Important)

When the error involves multiple objects, the selection of the reset\_object must respect **operation dependency**:

- If a target object is **blocked, pressed, clamped, or locked** by another, the first step of recovery should address the obstructing object.  
- Therefore, the reset\_object should be the **first object that needs to be manipulated** in order to recover from the failure, not necessarily the final target you want to restore.

**Example:**  
Error Group = [``coffee capsule", ``coffee machine"]
The coffee machine’s lid is closed (even if not completely), and the capsule is jammed inside the insertion slot.  
Although the ultimate goal is to reset the capsule, the first step is to **open the lid** — thus, the reset\_object is ``coffee machine".  

This rule ensures that the chosen reset action is **feasible and logically correct**, avoiding the selection of objects that are blocked or constrained.

---

\#\# Analysis Tasks

Please analyze the video and provide the following information:

\#\#\# 1. Failure Reason
- Describe in detail what caused the task to fail.  
- Explain why this failure occurred.  
- If multiple factors contributed, list them in order of importance.

\#\#\# 2. Failure Time Point
- Specify the exact time (frame index or timestamp) when the failure occurred.  
- Describe the scene state at that moment.  
- Explain why this moment represents the critical failure point.

\#\#\# 3. Recovery Strategy
Based on the failure scene state at that moment, determine the most appropriate recovery strategy:
- **retry** – for temporary issues (e.g., unstable grasp, small trajectory deviation) that can be resolved by retrying the same action.  
- **reset** – for state-altering issues (e.g., toppled cup, displaced tool) that require resetting one or more objects to their correct states.

\#\#\# 4. Reset Object (only if the strategy is reset)
Based on the failure scene state at that moment, find the object that needs to be reset:
- Specify which object should be reset (**choose only one**).  
- Describe **from what state to what state** it should be reset.  
- The object must appear in the **Resettable Objects** list.  
- If the error group has operation dependencies, select the object that should be **manipulated first** during recovery.  
  For instance, if a coffee capsule is trapped under the lid, the first step is to open the lid, so the reset\_object is ``coffee machine".

\#\#\# 5. Error Group
- Identify all objects involved in forming the error state.  
- List them in an array.  
- Ensure the reset\_object is included in this list.  
- If only one object is involved, the list should contain just that object.

---

\#\# Output Format
Please output your analysis in the following **JSON** format:

```json

\{

  ``failure\_reason": ``Detailed description of the failure cause.",

  ``recovery\_strategy": ``retry or reset",

  ``reset\_object": ``Name of the object to reset (null if strategy is retry)",

  ``error\_group": [``List of objects involved in the error; must include reset\_object if not null"],

  ``failure\_timestep": 1234,

  ``failure\_description": ``Detailed description of the failure moment.",

  ``confidence": 0.8

\}
'''
\end{tcolorbox}
\end{table*}

\begin{table*}[t] 
    \centering
\caption{The main prompt used in our experiment for failure case detection during the closed-loop inference for ThreePieceAssembly.}
\label{tab:prompt_for_closed_loop_inference}
\begin{tcolorbox}[title=Failure Detection Query, width=\textwidth]
  \footnotesize
You are an expert in analyzing robot task execution scenes, skilled at identifying abnormal states and suggesting appropriate recovery strategies.

The current task is called **Three Piece Assembly**, and its goal is to “assemble the three pieces.”  

The correct execution sequence is as follows:

1. Place the **T-shaped block** into the **square block**.  

2. Then place the **U-shaped block** on top of the **T-shaped block**.

The following image shows the current state of the task:

Please analyze the current scene.  
If the **T-shaped block** is visible and not placed flat when inserted into the **square block** — for example, it is tilted or stuck — the answer is **B**.  
If the **U-shaped block** is not placed flat when put on the **T-shaped block** or is lying on the desk, the answer is **C**.
Otherwise, the answer is **A**.  

At the end of your answer, clearly state your conclusion in the following format:  
**The answer is [A] / [B] / [C].**
\end{tcolorbox}
\end{table*}

\section{Broader Impact and Future Work}

\noindent \textbf{Broader Impact} The core value of the FLARE framework lies in endowing robotic systems with higher autonomy and reliability, which is crucial for advancing VLA models from lab demonstrations to large-scale, dependable deployment in real-world applications.

By significantly reducing the need for human intervention, FLARE can boost the operational efficiency and safety of industrial automation, service robotics, and professional service domains. This increase in robustness is a critical step in the transition of robots from simple assistive tools to reliable, persistent collaborators, particularly in environments that are repetitive, high-risk, or difficult to access.

As with any highly autonomous system, the powerful self-correction capabilities of FLARE must be accompanied by careful safety and ethical protocols. We therefore emphasize the following:

All recovery and reset actions must strictly adhere to pre-defined safety constraints to ensure they do not cause unintended harm to the environment or personnel.

We must maintain the system’s interpretability and transparency, and ensure that operators retain an emergency override or control switch to guarantee human oversight and accountability remain the final authority over the system.

\vspace{2mm}
\noindent \textbf{Future Work}
While the FLARE framework has markedly advanced VLA models' capabilities in failure recovery, there remains substantial room for exploration in terms of scalability and generalization. Our future work will focus on the following three main directions:

\begin{enumerate}
    \item Towards Generative and Generalized Recovery Policies: Our current robust ``Reset" mechanism relies on a pre-defined library of recovery skills, which limits its ability to handle Out-of-Distribution (OOD) and complex failure states that require non-standard or multi-step, collaborative actions. Future work will explore leveraging the powerful reasoning capabilities of large Vision-Language Models (VLMs) to enable them to autonomously infer and generate custom, ad-hoc recovery action sequences based on the current failure state and the desired goal. Achieving an end-to-end generative recovery policy is a key step in boosting FLARE’s generalization.

    \item Integrating Online Adaptation and Lifelong Learning: To address the data collection and annotation challenges associated with scaling our current approach, we plan to integrate the FLARE framework into an Active Learning loop. When the model encounters an unseen failure mode with low confidence, the system will be able to automatically trigger data collection and policy fine-tuning in a controlled environment. This mechanism for continuous learning will allow the deployed robot to persistently enhance its own recovery capabilities throughout its operational lifespan.

    \item Transfer Validation for Real-World Complexity: Our ultimate goal is to validate FLARE's effectiveness in more challenging, real-world scenarios. This includes testing the system's robustness in environments with dynamic changes, when handling non-rigid or deformable objects, and during long-horizon tasks that require human collaboration. These practical validations will drive the transition of FLARE from a laboratory framework to a real-world utility.
\end{enumerate}

\end{document}